\documentclass[letterpaper, 10 pt, conference]{ieeeconf}  

\IEEEoverridecommandlockouts                              

\let\labelindent\relax  

\usepackage{cite}
\usepackage{gatherenum}
\usepackage{amsmath,amssymb,amsfonts}

\usepackage{graphicx}
\usepackage{textcomp}
\usepackage[hidelinks]{hyperref}
\usepackage{authblk}
\usepackage{url}
\usepackage{color}
\usepackage{colortbl}
\definecolor{Ocean}{RGB}{186, 222, 244}
\usepackage{graphicx}
\usepackage{gensymb}
\usepackage{algorithmic}
\usepackage{algorithm}

\usepackage[utf8]{inputenc}
\usepackage[T1]{fontenc}
\usepackage[ngerman]{babel}

\usepackage{booktabs}
\usepackage{multirow}
\usepackage[table,xcdraw]{xcolor}

\usepackage{etoolbox}
\makeatletter
\patchcmd{\@makecaption}
{\scshape}
{}
{}
{}
\makeatother

\title{\LARGE \bf
	SCALE: Self-Correcting Visual Navigation for Mobile Robots via Anti-Novelty Estimation}

\author{Chang Chen$^{1}$, Yuecheng Liu$^{2}$, Yuzheng Zhuang$^{2}$, Sitong Mao$^{3}$, Kaiwen Xue$^{3}$ and Shunbo Zhou$^{3,\dagger}$
	\thanks{$^{1}$ Work done during an internship in Huawei. School of Science and Engineering,
		The Chinese University of Hong Kong, Shenzhen.
		{\tt\small changchen@link.cuhk.edu.cn}}%
	\thanks{$^{2}$ Huawei Noah's Ark Lab.
		{\tt\small liuyuecheng1, zhuangyuzheng@huawei.com}}%
	\thanks{$^{3}$ Edge Cloud Innovation Lab, 
		Shenzhen Huawei Cloud Computing Technologies Co., Ltd. {\tt\small maositong, xuekaiwen, zhoushunbo@huawei.com}}%
	
	\thanks{$^{\dagger}$ Shunbo Zhou is the corresponding author.}
}

\begin{document}
	
	\maketitle
	\thispagestyle{empty}
	\pagestyle{empty}

	\begin{abstract}
		Although visual navigation has been extensively studied using deep reinforcement learning, online learning for real-world robots remains a challenging task. Recent work directly learned from offline dataset to achieve broader generalization in the real-world tasks, which, however, faces the out-of-distribution (OOD) issue and potential robot localization failures in a given map for unseen observation. This significantly drops the success rates and even induces collision. In this paper, we present a self-correcting visual navigation method, SCALE, that can autonomously prevent the robot from the OOD situations without human intervention. Specifically, we develop an image-goal conditioned offline reinforcement learning method based on implicit Q-learning (IQL). When facing OOD observation, our novel localization recovery method generates the potential future trajectories by learning from the navigation affordance, and estimates the future novelty via random network distillation (RND). A tailored cost function searches for the candidates with the least novelty that can lead the robot to the familiar places. We collect offline data and conduct evaluation experiments in three real-world urban scenarios. Experiment results show that SCALE outperforms the previous state-of-the-art methods for open-world navigation with a unique capability of localization recovery, significantly reducing the need for human intervention. Code is available at
		\href{https://github.com/KubeEdge4Robotics/ScaleNav}{\textcolor{cyan}{https://github.com/KubeEdge4Robotics/ScaleNav}}.
		
	\end{abstract}
	
	\section{INTRODUCTION}
	Learning-based visual navigation algorithms have been extensively studied in recent years. However, most of the prior studies explore the algorithms in simulation environment that posit extensive online interaction and often suffer from limited robustness and generalization when transferring to the real world due to the sim-to-real gap. Recently, there is some work dedicated to lift the assumptions by learning directly from offline real-world data that does not require collection from experts\cite{sptm2018, ving2021, shah2022offline, shah2023vint}, which is encapsulated as ``experience learning'' in \cite{levine2023experience}. It has been proven to be a promising direction to achieve broader generalization for executing real-world robot tasks. Among these offline learning methods, a key component is the goal-conditioned value estimation that indicates the traversability from current state to goal by predicting the expected number of timesteps required to transit. It can be efficiently learned by either imitation learning\cite{sptm2018, ving2021,shah2023vint} or offline reinforcement learning\cite{shah2022offline}, while the latter does not necessitate the expert data and can control the learned policies' preference through reward design. By building an image-node topological graph in the deployed environment, this value estimation can also act as a similarity measurement for retrieval-based visual localization when executing long-horizon tasks in the region repeatedly. 
	
	\begin{figure}[t]
		\centering
		\setlength{\abovecaptionskip}{-0.4cm}
		\includegraphics[width=0.48\textwidth]{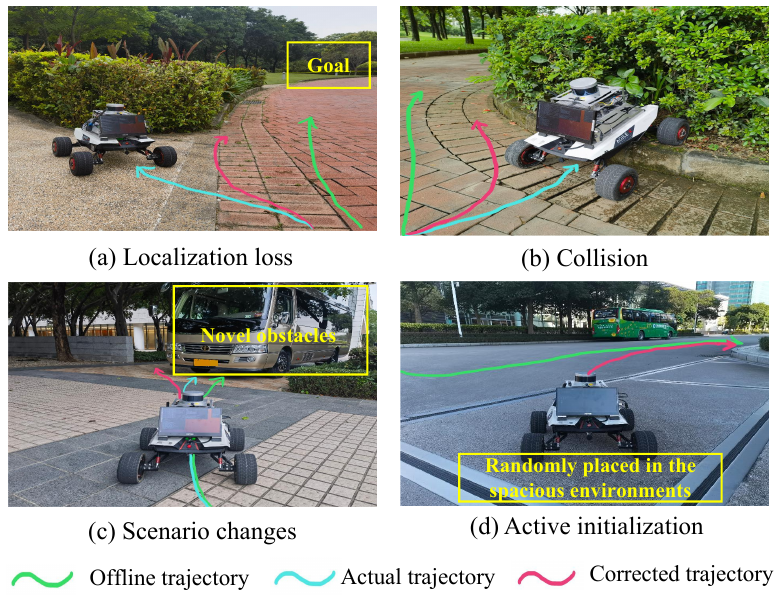}
		\caption{\textbf{Localization challenges}. For the offline learning methods, the out-of-distribution issue frequently arises in the real-world navigation, making the robot lose its localization in the given map that decreases the success rates (a) and even causes collision (b). When the scenario changes that some novel obstacles that are not in the offline dataset appear during deploying, the policy may also become unavailable (c). Another case is when the robot is placed at an unknown position in the spacious deploying environments (d). A novel localization recovery module can tackle these four typical cases by predicting future states and assessing their novelty.
		}\label{correction}
	\end{figure}
	
	However, due to the limited state distribution in the offline dataset, when facing the  illumination and scenario changes and the accumulative error during navigation process, the localization loss often arises that the robot cannot localize itself on a given map. Despite these perturbation to vision encoder can be alleviated by adopting vision augmentations such as semantic segmentation and random masking during training, the robust policy towards limited state distribution still remains a challenging issue when deploying offline learning methods to real-world tasks. In this case, the value estimation of current observation and the desired goal becomes too low leading to low confidence of the predicted policy. Hence, the robot may make infeasible decision, which significantly decrease the success rates and even result in collision that require tedious human monitoring and correction.

	In fact, the localization loss issue was discussed earlier in \cite{fan2019unlost} that considered the environments with dense crowd. It detected some recovery points on the grid map and evaluated their accessibility using a value function to get the robot unlost, but required online training and LIDAR points for SLAM. Another approach opts to learn from the offline experience to predict whether disengagement, i.e., human intervention, will happen in order to avoid them\cite{kahn2021land}, but requires collecting the experience of disengagement at test time. Motivated by this realistic issue, we pose the question: ``can we leverage the prior knowledge from offline data to guide the robot to correct its trajectory autonomously without human intervention and specific data collection?'' (see fig. \ref{correction})

	In this paper, we study how to address the aforementioned issue under a scalable condition: only forward-facing RGB camera is accessible during inference. Our key insight is that, instead of learning from the experienced events to predict whether the localization loss or collision will happen, we predict whether the future states lie in the offline data distribution, i.e., ``anti-novelty''. To this end, we build upon prior work and propose SCALE, a novel \textbf{S}elf-\textbf{C}orrecting visual navigation method via \textbf{A}nti-nove\textbf{L}ty \textbf{E}stimation. We use an affordance model to generate potential future trajectories and plan on these trajectories under the guidance of novelty predictor learned by random network distillation (RND)\cite{burda2018exploration}. The optimal candidate minimizing the cost function can induce the robot to the familiar places with the least novelty.

	The main contributions of this paper are summarized as: 1) We develop an image goal-conditioned visual navigation without LIDAR or GPS based on offline reinforcement learning method, implicit Q-learning (IQL)\cite{kostrikov2021offline}. 2) We propose a novel localization recovery method. We self-supervisedly learn an affordance model to generate multi-step future trajectories, and learn a novelty predictor via RND to estimate the future novelty of these trajectories The candidate with the least novelty is selected to guide the localization recovery. 3) Extensive real-world experiments are conducted and compared with state-of-the-art baselines. Results demonstrates that the proposed method significantly outperforms the baselines.
	

	\section{Related Work}
	\subsection{Visual Navigation by Offline Learning} \label{topological}
	Prior work has studied learning-based visual navigation algorithms and proven scene memory or map to be the key to handle long-horizon visual navigation. Metric map based memory projects the visual input to the top-down occupancy grid map\cite{parisotto2018neuralmap, henriques2018mapnet, li2019drlexplore, chaplot2020learning, min2022film}, but assume to access accurate LIDAR or depth for map construction. Non-parametric topological map, on the other hand, efficiently compresses the scenario by some key frames. The combination of topological map and agent's policy can perform in either end-to-end\cite{fang2019scene, vgm2021, kim2022topological}, or modular design\cite{sptm2018, ntslam2020}. However, they require the agents to train from online interaction in simulation, and suffer from sim-to-real gap when deploying to the real world.
	
	To lift the restrictive assumptions, some recent work focuses on learning the geometric attributes of environments directly from offline real-world data\cite{hahn2021nrns, gregory2021badgr, ving2021, shah2022offline, shah2023vint}. 
	ViNG\cite{ving2021} learns an inverse dynamic model to predict the temporal distance and relative poses between pair images from offline data. ReViND\cite{shah2022offline} adopts IQL to learn the value function and customized policy, which are combined with a topological map to break down long-horizon tasks to some simple subtasks. Nevertheless, they are often troubled by localization loss issue when deploying to the real world.
	

	\subsection{Generalization to Out-of-Distribution}
	Prior work has addressed the out-of-distribution (OOD) issue of offline learning methods by either collecting disengagement dataset at test time and learn to predict and avoid the disengagement\cite{kahn2021land}, or using the prior knowledge from offline dataset to accelerate exploration and fine-tuning on novel environments\cite{shah2021rapid, khazatsky2021val, fang2022ptp, fang2022flap, bharadhwaj2023visafford}. Specifically, RECON\cite{shah2021rapid} adopts a variational information bottleneck (VIB)\cite{alemi2021vib} to learn a compact latent goal distribution and navigate to the latent goal when it is feasible otherwise requiring uninformed frontier exploration. FLAP \cite{fang2022flap} constructs a lossy representation space for generating some subgoals acting as the anchors for online fine-tuning on the novel manipulation tasks. 
	
	
	
	
	\begin{figure*}[t]
		\setlength{\abovecaptionskip}{-0.2cm}
		\centering
		\includegraphics[width=0.95\textwidth]{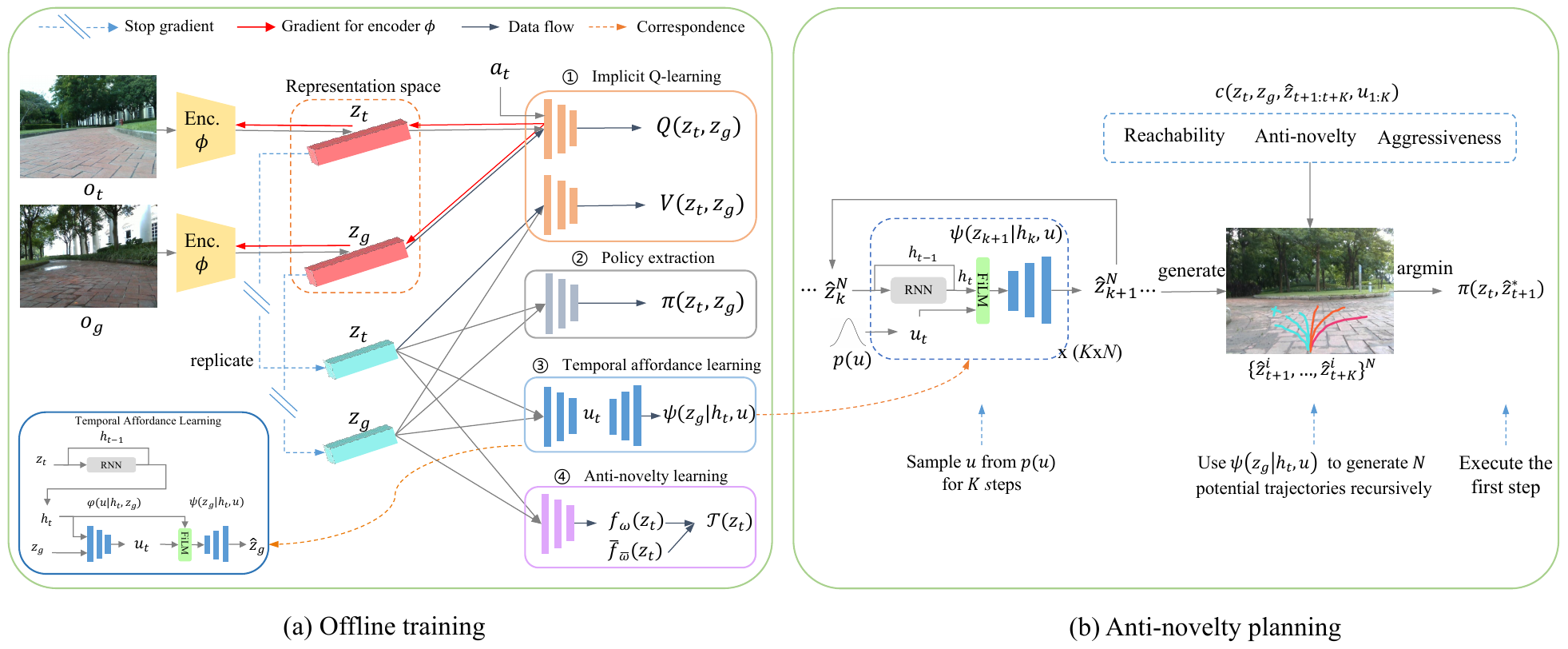}  
		\caption{\textbf{Overall framework}. (a) SCALE first pretrains a self-consistent representation space $z$ by the VAE-style loss, then fine-tunes it by the gradients from $Q (s, a, g)$ in the IQL. Next, we train the policy network $\pi$, temporal affordance model $\psi$ and the novelty predictor $f_\omega$ over the trained representation space. (b) When the robot gets lost, SCALE randomly samples some transition $u$ from the prior $p(u)$ and feds it to the temporal affordance model to generate some multi-step latent trajectories recursively. Then it evaluates the candidates in terms of the reachability, anti-novelty and aggressiveness. Finally, it selects the optimal trajectory and executes the first step, then repeats until being successfully localized again.}\label{framework}
	\end{figure*}
	
	\subsection{Problem Formulation} \label{reward}
	We consider an infinite-horizon goal-conditioned Markov Decision Process, with states $s \in \mathcal{S}$ referring to image observation, goals $g\in \mathcal{G}$ being presented in images instead of GPS positions, actions $a\in \mathcal{A}$ corresponding to the continuous linear and angular velocities of the mobile robot, reward function $r(s, a, g)$, environment dynamics $p(s^\prime|s, a)$ and the discounted factor $\gamma$. For the offline RL, an offline dataset $\mathcal{D}_\beta=\{(s,a,r,s^\prime) \}$ is pre-collected by a behavior policy $\pi_\beta$. The general objective of a goal-conditioned agent is to learn a policy $\pi(\cdot|s, g)$ that maximizes the expected cumulative discounted returns over the goal distribution $J(\pi)=\mathbb{E}_{a_t \sim \pi(\cdot|s_t, g), g \sim p_g} [\sum_{t} \gamma^t r(s_t, a_t, g)].$
	The reward function $r(s, a, g)$ is set to $-1$ for all steps except for $0$ for reaching goal and a penalty $-10$ for collision. 
	
	

	\section{METHODOLOGY}

	\subsection{Image-Goal Navigation by Implicit Q-Learning} \label{IQL}
	We adapt IQL, an offline reinforcement learning method that does not ever query the values of any unseen actions, to learn the image-goal conditioned policy and its corresponding value function. The IQL agent uses three networks to approximate the $Q$, target $Q$, and value function, respectively.
	IQL updates the $Q$-function $Q (s, a, g)$ by a SARSA-style objective with estimating the maximum $Q$-value: 
	\begin{equation} \label{qf}
		\begin{aligned}
			\mathcal{L}_Q &= \mathbb{E}_{(s,a,s^\prime)\sim \mathcal{D}, g\sim p(g|s)}\\
			&[(r(s,a, g)+ \gamma V(s^\prime, g)
			-Q (s,a, g))^2],  
		\end{aligned}
	\end{equation}
	where $V(s^\prime, g)$ is a value function used to approximate the target $Q$-function $\hat{Q} (s^\prime, a^\prime, g)$ to avoid the perturbation by some ``lucky'' samples that happened to transit into a good state. Since the $Q$-function cannot be trained on all possible actions by offline dataset, IQL employs an expectile regression $L_2^\tau$ to predict the upper expectiles of the distribution over $Q$-value as the approximation of the maximum $Q$-value. Thus, learning the value function yields the objective
	\begin{equation}\label{vf}
		\mathcal{L}_V=\mathbb{E}_{(s,a)\sim D, g\sim p(g|s)}[L_2^\tau(\hat{Q}(s, a, g)-V(s,g))],
	\end{equation}
	where $L_2^\tau(u)=|\tau-\mathbb{I}(u>0)|u^2$.
	Lastly, advantaged weighted regression\cite{peng2019advantageweighted} is used to extract the policy $\pi(s, g)$ from the $Q$-function with an inverse temperature $\beta$:
	\begin{equation}\label{pi}
		\begin{aligned}	
			\mathcal{L}_\pi&=\mathbb{E}_{(s,a)\sim D, g\sim p(g|s)} \\
			&[\exp(\beta(\hat{Q} (s,a,g)-V (s,g))) \log \pi (a|s,g)].
		\end{aligned}
	\end{equation}
	
	The IQL framework is illustrated in Fig. \ref{framework}. We consider image-goal rather than point-goal setting for its wide application in embodied tasks and can be used in GPS-denied environments. In practice, we find two techniques crucial for successfully training the image-goal conditioned IQL agent: 1) \textit{negative sampling} is necessary for value learning with paired image inputs. Similar as that in \cite{ving2021} but with different implementation, we assign image pairs that are separated less than or equal to a threshold of timesteps $d_\text{max}$ as the positive samples $B_+$, which update the value function by Eq. \ref{vf}. Those separated beyond the threshold are assigned as the negative samples $B_-$, which require the value function to predict a minimum threshold $V_\text{min}$. As such, the representation distance can be pushed away by the negative samples, improving the efficiency for learning the value prediction. 
	2) \textit{relative goal embedding}, the difference between goal and current image embeddings, i.e., $\Delta z_{g,t}=z_g - z_t$, is more effective than directly using goal embedding $z_g$ as the input for the goal-conditioned networks to perceive the goal orientation.

	
	\subsection{Trajectory Generation by Affordance Learning} \label{affordance}
	
	Under an OOD situation where the predicted value falls below a threshold, the policy prediction will become unattainable to reach the goal. In this case, we use an affordance model\cite{gibson1977affordance} to generate a set of potential future trajectories that are reasonable and reachable, then select the optimal candidate that minimizes the cost function to help robot recover the localization autonomously.
	A general visual affordance model $p(z^\prime|z)$ can be represented by a generative model, which is trained to maximize $\mathbb{E}_{(z,z^\prime)\sim \mathcal{D}} [\log p(z^\prime|z)]$, where $z^\prime$ denotes the goal embedding for exploring in novel environments under current state $z$.
	Nevertheless, since the observation losing the localization can be greatly diverge from the goal, directly sampling a viable final subgoal from the goal distribution is often difficult\cite{khazatsky2021val, shah2021rapid}. Inspired by \cite{fang2022flap} that combines affordance learning with the latent goal-conditioned planning\cite{nasiriany2019planning} to break down the long-horizon tasks into some feasible subtasks, we construct a conditional affordance model for multi-step trajectory imagination.
	
	We train the conditional affordance model $\psi(z^\prime|z, u)$ over the representation space without predicting the details of environment. To learn the self-consistent affordance, we additionally use an encoder $\varphi(u|z, z^\prime)$ to self-supervisedly capture the transition $u$ between current and goal state. As such, the paradigm completes a forward-inverse cycle consistency (FICC)\cite{ye2023become} to learn the representation of $u$ (see Fig. \ref{framework}a). In this paradigm there is a shortcut that $u$ can contain most information of $z_g$ and disregard the difference between $z$ and $z_g$, achieving zero loss in the cycle consistency but being meaningless. To avoid the shortcut, both VIB and vector quantization\cite{oord2018neural} can be useful, and we use VIB to regularize $u$ to preserve the minimal necessary information for its simplicity and efficiency. As such, the affordance model can be trained by minimizing the objectives
	\begin{equation}
		\begin{aligned}\label{afford}
			\mathcal{L}_{\text{afford}}&= -\mathbb{E}_{(z,z^\prime) \sim \phi} \log p(z^\prime|z, u)\\
			&  + \beta \mathbb{E}_{(z, z^\prime) \sim \phi } \left[D_{KL}(q(u|z, z^\prime) || p(u)) \right],
		\end{aligned}
	\end{equation}
	where $\phi$ is the vision encoder that will be discussed in \ref{implement}, the first term minimizes the affordance prediction loss and the second term prevents the shortcut.
	
	Furthermore, to facilitate predicting the forward dynamics in the decoder $\psi(z^\prime|z, u)$, we use feature-wise linear modulation (FiLM)\cite{ethan2021film} operation to condition the latent random code $u$ on current observation $z$ instead of directly concatenating the two terms:
	\begin{equation}
		f(z, u) = \gamma(z) \odot u + \delta(z),
	\end{equation}
	where $\gamma$ and $\delta$ are two linear layers.
	

	
	\begin{figure}[t]
		\centering
		\setlength{\abovecaptionskip}{-0.3cm}
		\includegraphics[width=0.48\textwidth]{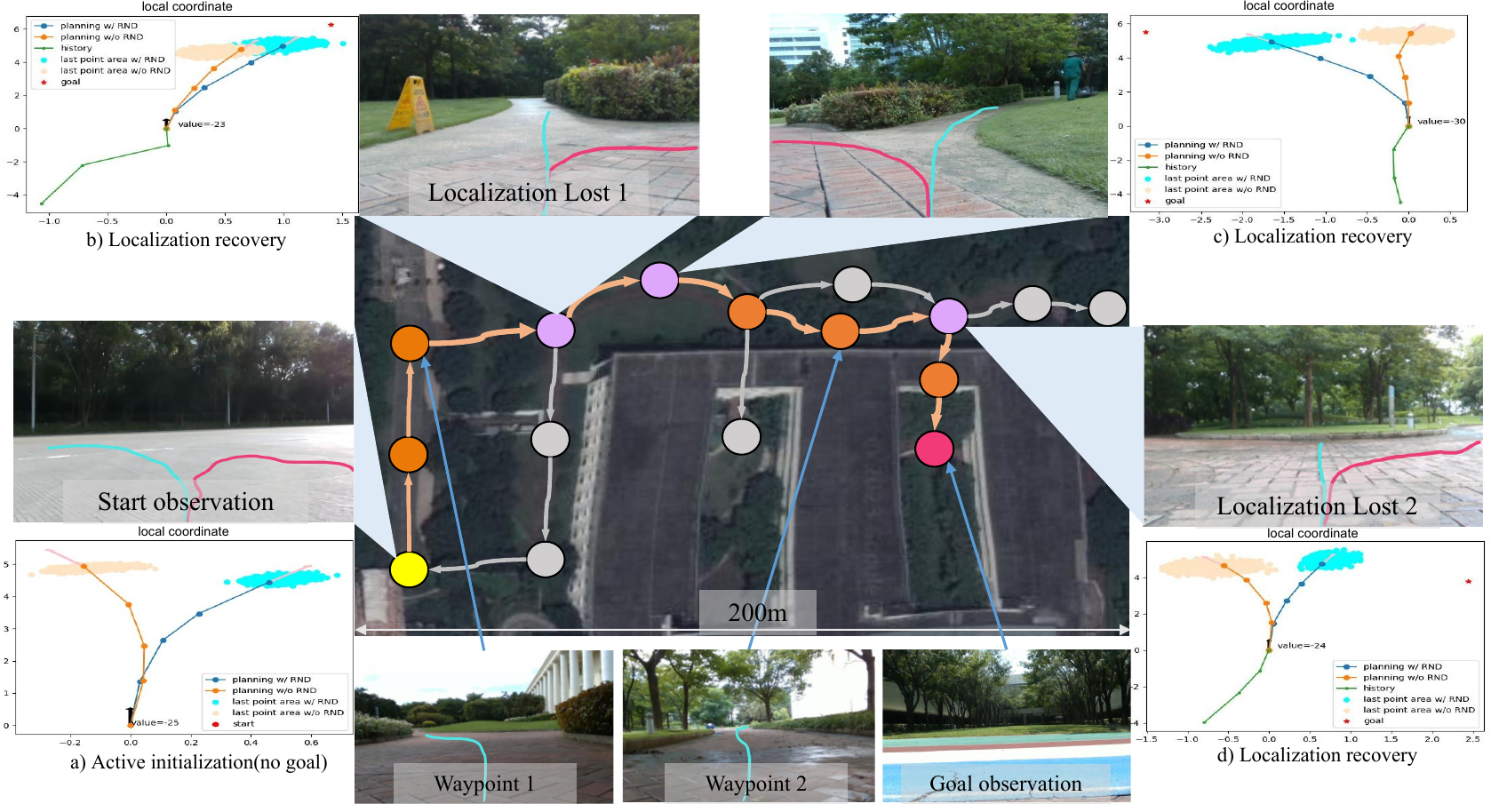}
		\caption{\textbf{Topological navigation with localization recovery}. SCALE combines the topological visual navigation with a novel localization recovery module. We first build a topological map (gray cycles and lines) based on the offline dataset. Next, starting at the yellow cycle, we use the localization module to do active initialization. Then, given a goal image, we search a route (orange cycles) on the topological graph and execute to the goal (red cycle) step by step. The cyan lines denote the actual trajectories. The plot panels show the plans with and without RND for the active initialization and localization recovery (purple cycles) during navigation. Ultimately, the optimal trajectories (red lines) guide the robot to relocalize itself.}\label{procedure}
	\end{figure}
	
	\subsection{Aggressive Prediction by Temporal Features}
	Utilizing a single front-facing observation to envision a viable subgoal for navigation tasks can be intractable, particularly when the robot deviates so far that requires an aggressive view change, i.e., $\Delta \theta > \Delta \theta_{\text{min}}$, to relocalize itself. This stems from the fact that the environment is only partially observed, making it difficult to predict subgoals outside the current field-of-view. To address this issue, a key insight is the previously localized states can provide surrounding pixels for the agent to anticipate subgoal states outside the view accurately. Therefore, we use the localized history states to endow the aggressive prediction capability. Specifically, a recurrent neural network (RNN) is adopted to extract the temporal features $h_t$ from the stacked history encodings:
	\begin{equation}
		h_t = \text{RNN}(z_{t-H:t}, \theta),
	\end{equation}
	which is then fed to the affordance model as the condition:
	\begin{equation}
		u_t = \varphi(h_t, z_g), \ \hat{z}_g= \psi(h_t, u_t),
	\end{equation}
	where $h_t$ is the temporal encoding at step $t$ and $H$ is the history horizon. In practice, we use gate recurrent unit (GRU)\cite{chuang2014gru} as the efficient temporal encoder.


	\subsection{Anti-novelty Guidance by Random Network Distillation} \label{conservatism}
	
	To guide the robot to the familiar places, we estimate the future novelty at current state and guide the robot to the places with the least novelty. In contrast to the widely used exploration methods that seeking for novelty, we seek for ``anti-novelty''\cite{rezaeifar2021offline} that avoided selecting actions that could lead to unpredictable consequences. Here we seek for selecting the potential future states with the least novelty. There are several approaches to measure the novelty, such as 
	conditional VAE\cite{rezaeifar2021offline}, variance of deep ensemble\cite{lakshminarayanan2017ensemble},  and RND\cite{burda2018exploration, nikulin2023anti}. We adopt RND, a learning-based visitation count, to evaluate the novelty over the latent space, which is proven to be discriminative enough to novel input\cite{nikulin2023anti}.
	
	RND was originally introduced to provide efficient intrinsic rewards to incentivize the exploration of novel environment with pixel-level input. It consists of a randomly initialized prior network $\bar{f}_{\bar{\omega}}$ without gradient update, and a predictor network $f_\omega$ that learns to predict the output of prior network given the same input. Then the output difference between two networks indicates a novelty metric $\mathcal{T}(z)$:
	\begin{equation}\label{rnd}
		\mathcal{T} (z) = \lVert \bar{f}_{\bar{\omega}} (z) - f_\omega(z) \rVert _2^2.
	\end{equation}
	After training, the output of predictor network will be close to that of the prior with seen input, while being distinct to that of the prior with unseen input. By incorporating this metric into the cost function, we can select the trajectory candidate that approaches to the familiar places.

	\begin{algorithm}[t]
		\caption{Training SCALE}
		\label{training}
		\begin{algorithmic}[1]
			\STATE Initialize  dataset $\mathcal{D}$, $\phi(z|s)$, $Q(s, a, g)$, $\hat{Q}(s, a, g)$, $V(s, g)$, $\pi(s, g)$, $\varphi(u|z, z^\prime)$, $\psi(z^\prime|z, u)$,  $f_\omega (z)$, $\bar{f}_{\bar{\omega}}(z)$
			\STATE Pretrain $\phi(z|s)$ on $\mathcal{D}$ by $\mathcal{L}_\text{vqvae}$; \COMMENT{stage 1} 
			\FOR[stage 2] {each gradient step} 
			\STATE Sample batch $\mathcal{B}_+ \sim \mathcal{D}_+$, $\mathcal{B}_- \sim \mathcal{D}_-$;
			\STATE Update $Q (s, a, g)$, $V(s, g)$, $\pi(s, g)$, $\psi(z^\prime|z, u)$ \\
			by Eq. \ref{qf}, \ref{vf}, \ref{pi}, \ref{afford}, respectively;
			\STATE Update $\phi(z_t|s_t)$ by the gradient from $Q(s,a,g)$;
			\STATE Soft update $\hat{Q}(s, a, g)$ by $Q(s, a, g)$;
			\ENDFOR
			\STATE Train $f_\omega (z)$ with $\bar{f}_{\bar{\omega}}(z)$ on $\mathcal{D}$ by Eq. \ref{rnd} \COMMENT{stage 3} 
		\end{algorithmic}
	\end{algorithm}
	
	Consequently, we tailor the cost function as minimizing the novelty and optionally the representation distance to the goal, with constraints on the reachability, the aggressiveness, and probability that happening, which are written as
	\begin{equation}\label{cost function}
		\begin{aligned}
			c&(z_t, z_g, \hat{z}_{t+1:t+K}, u_{1:K}) = \mathcal{T} (\hat{z}_K) + \lambda  \lVert \hat{z}_K - z_g \rVert_2^2 \\
			& + \sum_{k=1}^{K} (\eta_1 (V_{\text{loc}} - V_k) + \eta_2 \log p(u_k)) 
			+ \eta_3  (\Delta \theta_{\text{min}} - \Delta \theta_K),
		\end{aligned}
	\end{equation}  
	where $\mathcal{T}(\hat{z}_K)$ denotes the novelty estimation of $\hat{z}_K$, $\lVert \hat{z}_K - z_g \rVert_2^2$ is an optional objective that minimizes the representation distance between the final state of the rollout and the goal if given. For constraints, $V_k$ denotes the value estimation $V(\hat{z}_{k-1}, \hat{z}_k)$,
	$p(u_k)$ denotes the probability of the transition $u_k$ that happening,  $\Delta \theta_K$ denotes the aggressiveness represented by the related yaw differing to the current state, $V_{\text{loc}}$ and $\Delta \theta_{\text{min}}$ are the minimum thresholds for localization and aggressiveness, and $\eta_{1:3}$ are the Lagrange multipliers. 
	
	We also use the model predictive path integral (MPPI)\cite{gandhi2021mppi} to iteratively improve the plans via importance sampling. Fig. \ref{procedure} shows a complete procedure of applying our localization recovery module in the topological visual navigation to improve the robustness towards the real-world challenges.

	\begin{figure}[h]
		\centering
		\setlength{\abovecaptionskip}{-0.4cm}
		\includegraphics[width=0.49\textwidth]{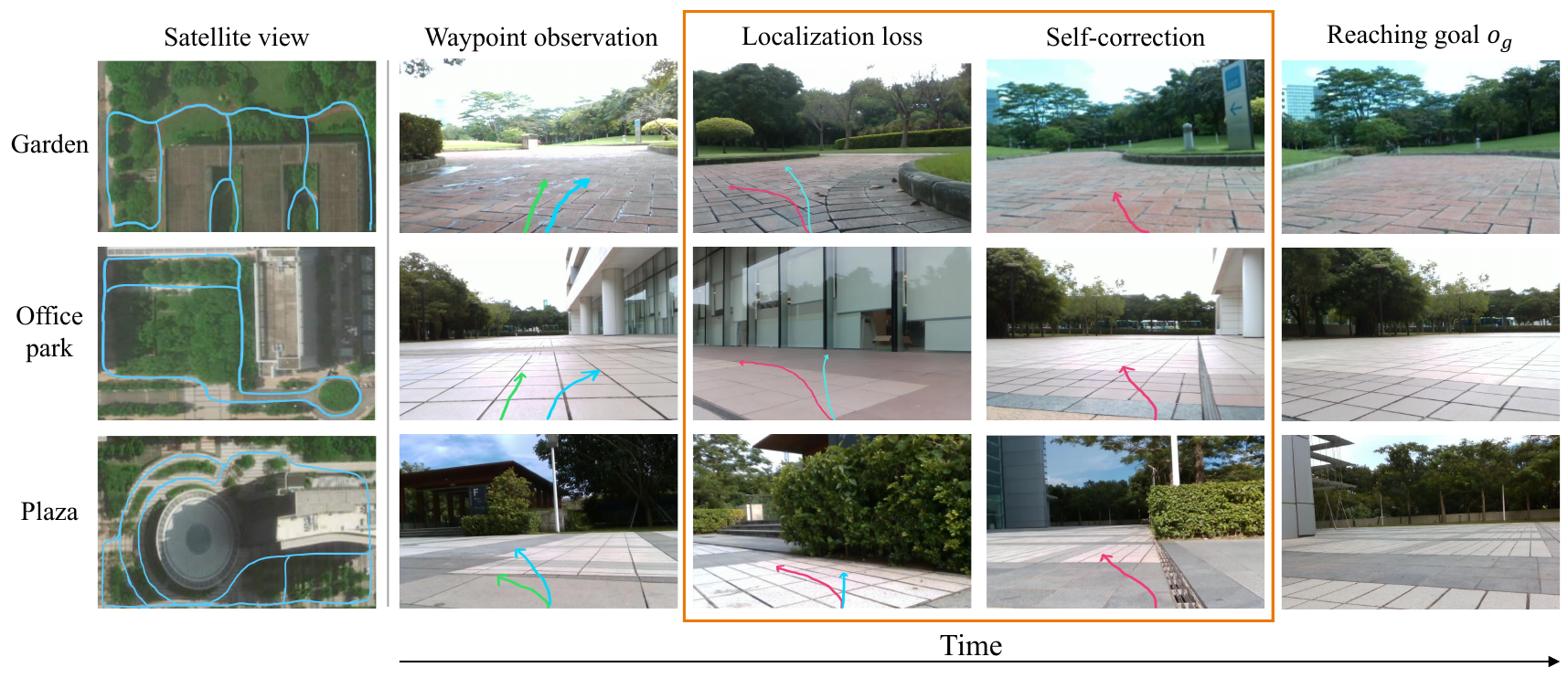}
		\caption{\textbf{Quantitative experiments}. We evaluate SCALE in three outdoor environments, which are shown in the satellite images (1st column) and the cyan lines indicate the navigation routes. The 2nd column shows some waypoints before the localization failures. When the actual trajectories (cyan lines) distinctly deviate from the topological map built on the offline trajectories (green lines), the localization failures arises (3rd column). In this case, our localization recovery module generates some latent subgoals and evaluates them by cost function. The optimal anti-novelty plan (red lines) is executed to correct the robot's trajectory (4th column), eventually navigating the robot to the goal (5th column) without human intervention.}\label{experiments}
		\vspace{-0.3cm}
	\end{figure}
	
	\section{Experiment}
	\subsection{Experiment Setup}
	\textit{1) Robot Platform:} We implement the algorithms on a Scout-mini robot (see Fig. \ref{correction}). The sensor suite that we use consists of a forward-facing Realsense D435 RGB camera with $69\degree$ field-of-view for the environment observation, and a wheel odometry for approximating pose estimation. Compute is provided by an NVIDIA Jetson Xavier computer. The 3D
	LIDAR on the robot is not used in this work.
	
	\textit{2) Data Collection:} We use the above robot to manually collect diverse training data $\mathcal{D}$ in three scenarios, including both expert and collision trajectories in the morning and afternoon. For each scenario we collect totally 35-minute trajectories. At each timestep the observation is an onboard $480 \times 640$ RGB image, and the estimated pose $(x, y, \theta)$ from odometry. Note that poses are only used to calculate the average speeds to train the policy network, while only image observation is necessary during deployment.
	
	\textit{3) Implementation Detail:}\label{implement} To facilitate learning the goal-conditioned policy, we use hindsight goal relabeling\cite{andrychowicz2018hindight} to improve learning the policy from sparse rewards. We first pretrain a Vector Quantization VAE (VQ-VAE)\cite{oord2018neural} by loss $\mathcal{L}_\text{vqvae}$ on the offline dataset to learn a self-consistent representation that is robust to the illumination changes across a day, then use its encoder as our vision encoder $\phi(z|s)$.
	We also use layer normalization\cite{ba2016layernorm} in the encoder to improve its efficiency. 
	$z_t$ and $\Delta z_{g,t}$ mentioned in section \ref{IQL} are then fed to $Q$, value, and policy network, respectively. The policy and affordance networks are simultaneously updated with the value network since they do not influence the value learning, while the novelty predictor $f_\omega (z)$ is trained after representation learning. We train the models with batch size of 128. We perform gradient updates by AdamW\cite{adamw2017} optimizer, with learning rate $\lambda=3\times10^{-4}$. We set expectile coefficient $\tau=0.7$ and the localization threshold $V_\text{loc}=-10$. Algorithms \ref{training} and \ref{deploy} summarize our approach in the training and deployment stages, respectively. 
	

	We build a non-parametric topological map $\mathcal{M}$ based on offline dataset as the scene memory for robot localization and long-horizon task planning, while this is not necessary for using SCALE. Following \cite{shah2022offline}, each node $v_j$ in the graph refers to an image, and each edge $e_{ij}$ represents the reachability, with a corresponding cost of $V(z_{v_i}, z_{v_j})$ whose absolute value refers to the expected discounted number of steps required for reaching $v_j$ from $v_i$. The localization loss refers to the predicted values between current observation and all nodes in the map are lower than a threshold $V_\text{loc}$.

	%

	\begin{algorithm}[t]
		\caption{Deploying SCALE}
		\label{deploy}
		\begin{algorithmic}[1]
			\REQUIRE $\phi(z|s)$, $V(s, g)$, $\pi(s, g)$, $\psi(z^\prime|z, u)$,  $f_\omega (z)$, $\bar{f}_{\bar{\omega}}(z)$
			\REQUIRE $o_t$, (optional) $o_g$, topological map $\mathcal{M}$
			
			\STATE $z_t \leftarrow \phi(o_t)$; $z_g \leftarrow \phi(o_g)$ if $o_g$ given;
			\WHILE[localization]{$ \max \{V(z_t, z_{v_m})\}^\mathcal{M} < V_\text{loc}$}
			\STATE Sample $u_{1:K}^N$ from $p(u)$ and generate multi-\\
			step trajectories $\hat{z}_{t+1:t+K}^N$  by $\psi$ recursively;
			\STATE Estimate $c(\cdot)$ in Eq. \ref{cost function} by $V$, $f_\omega$ and optional $z_g$;
			\STATE Select optimal trajectory $\hat{z}_{t+1:t+K}^*=\arg\min c(\cdot)$;
			\STATE Execute the first step of plan $\hat{z}_{t+1}^*$ by $\pi(z_t, \hat{z}_{t+1}^*)$; 
			\STATE Observe new $o_{t}$; $z_t \leftarrow \phi(o_t)$;
			\ENDWHILE
			\STATE $o_t$ can be localized to node $v_j = \arg \max \{V(z_t, z_{v_m})\}^\mathcal{M}$ on $\mathcal{M}$;
			\IF[navigation]{$o_g$ is given}
			\STATE Localize $o_g$ on $\mathcal{M}$;
			\STATE Search a route $\{v_j, \dots, v_g\}$ to reach goal $o_g$;
			\STATE Navigate to next waypoint $v_{j+1}$ by $\pi(z_t, z_{v_{j+1}})$;
			\ENDIF
			
				%
				%
			
			\end{algorithmic}
		\end{algorithm}
		
		\begin{table*}[t]
			\begin{center}
				\caption{\textbf{Success rates of image-goal navigation.} We evaluate the methods in three scenarios with three difficulties (Easy: <50m, medium: 50-150m, hard: 150-200m). There are some dynamic shuttle buses and sparse crowds in the office park and plaza scenarios.}
				\resizebox{1.0\linewidth}{!}{
					\begin{tabular}{cccccccccccc}
						\toprule
						
						Methods &\multicolumn{3}{c}{Garden \scriptsize{(270x50m)}} &\multicolumn{3}{c}{Office park \scriptsize{(150x100m)}} &\multicolumn{3}{c}{Plaza \scriptsize{(200x150m)}} & \multirow{2}{*}{\shortstack{Avg.\\ success rates}} 
						& \multirow{2}{*}{\shortstack{Avg. distance  until \\intervention(meters)}}\\
						\cmidrule(r){2-4} \cmidrule(r){5-7}  \cmidrule(r){8-10} 
						
						& Easy & Medium & Hard & Easy & Medium & Hard &Easy & Medium & Hard & &\\
						\midrule
						ViNG\cite{ving2021}              & 8/10 &6/10 &5/10 & 8/10 &5/10  &5/10 & 8/10 &5/10 &3/10 & 0.59 & 82.7\\   
						
						ReViND\cite{shah2022offline}            & 8/10 &5/10 &5/10 & 8/10 &6/10 &5/10 & 8/10 &6/10 &4/10 & 0.61 & 73.1\\ 
						SCALE w/o recovery (Ours) & 8/10 &7/10 &6/10 & 9/10 &8/10 &6/10 & 8/10 &6/10 &5/10 & 0.70 &102.4\\ 
						\rowcolor{Ocean} SCALE w/ recovery (Ours) & \textbf{9/10} &\textbf{8/10} &\textbf{8/10} & \textbf{10/10} &\textbf{9/10} &\textbf{8/10} & \textbf{9/10} &\textbf{8/10} &\textbf{8/10} &\textbf{0.86} &\textbf{160.3}\\ 
						
						\bottomrule
					\end{tabular}
				}
				\label{tab1}
			\end{center}
			\vspace{-0.3cm}
		\end{table*}

		\subsection{Performance comparison}
		
		We compare SCALE with and without recovery module to the following two state-of-the-art baselines:
		
		
		\textit{ViNG}\cite{ving2021}: A method that learns an inverse dynamic model to predict temporal distance and relative poses between any two images from the offline trajectories, and use a topological graph for high-level planning.
		
		\textit{ReViND}\cite{shah2022offline}: A method that applies the IQL to visual navigation and combines a topological map for long-horizon planning. For fair comparison, we also use image goal without accessing to GPS, and use the negative sampling strategy to enable learning the $Q$-function.
		
		

\begin{figure}[t]
			\centering
			\vspace{-0.2cm}
			\includegraphics[width=0.46\textwidth]{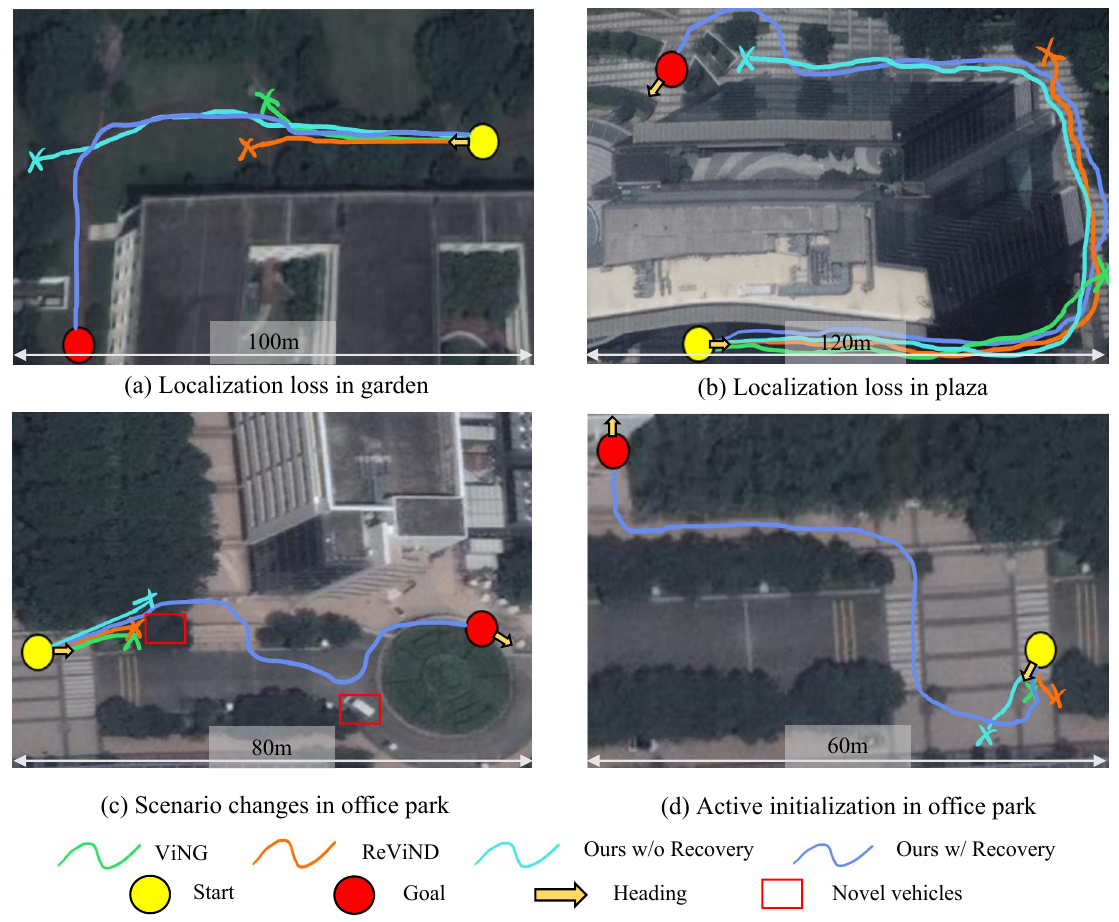}
			\setlength{\abovecaptionskip}{-0.2cm}
			\caption{\textbf{Performance demonstration.} Only SCALE equipped with localization recovery successfully reaches the designated goal, exhibiting strong robustness to the trajectory deviation induced by cumulative driving error (a), and the sharp turns through aggressive state prediction (b). SCALE uniquely succeeds in attaining the goal against the scenario changes (c) and active initialization in an unknown place in the spacious environments (d), which other three methods cannot handle.}\label{trajectories}
			\vspace{-0.45cm}
		\end{figure}

		We train the models on our collected offline dataset and evaluate the performance in terms of the image-goal navigation in three urban scenarios: garden, office park and plaza, with three difficulties: $<50m$, $50-150m$ and $150-200m$. For each difficulty we conduct $10$ trials. Fig. \ref{experiments} demonstrates the quantitative experiments of SCALE in these scenarios. And the success rates of image-goal navigation are provided in Table \ref{tab1}. 
		We see that ViNG and ReViND have comparable success rates while ViNG achieves slightly longer average distances until intervention than ReViND, since ViNG primarily follows the offline trajectories that can navigate farther when no localization loss happens. However, both often fail in the hard scenario (plaza) and the long-horizon trials (150-200 meters), which correspond to the localization loss and collision due to the cumulative error, scenario changes (additional parked or left cars), and infeasible policy. Compared with ReViND, SCALE without localization recovery can learn better policies from IQL agent which we think is due to the encoder pretraining and relative goal embedding, whereas it is still unable to handle the OOD situations in real-world tasks. By introducing the novel localization recovery module, the robustness towards the hard trials is significantly improved that SCALE with recovery can navigate x2 the distance until intervention than those of the two baselines with 26\% higher navigation success rate of than that of ReViND. Fig. \ref{trajectories} shows the performance of robustness towards the OOD situations of different systems, which proves that SCALE outperforms the two state-of-the-art baselines by tackling most of the real-world failure cases and saving the need for human monitoring and intervention. 
		
		Note that compared to SLAM-based methods, SCALE does not build a metric map but instead a topological map, thus requires lower device cost and enjoys faster inference speed for mapping and localization. Furthermore, our learning-based approach can adapt to the dynamic changes and be transferred to novel environments rapidly. 
		
		\subsection{Ablation Study}
		Learning an effective affordance model is the key to SCALE. Therefore, we design the ablation experiments to verify whether our chosen techniques in the affordance model can obtain gain, respectively. We compare SCALE to four ablations that are not using affordance, using affordance without RNN and RND, using affordance without RNN, and using affordance without RND. The average navigation success rates in three evaluation scenarios are shown in Table \ref{ablation2}. We observe that without affordance model the trained IQL agent can often fail during navigation especially in the medium and hard trials. By introducing a vanilla affordance model, the agent can be able to search several potential future trajectories whereas often unavailable when the localization is lost. On the other hand, RNN can help preserve as less as possible information in $u$ that promotes the accuracy of aggressive affordance prediction during inference, particularly when the robot drives into a corner. And SCALE employing the anti-novelty guidance learned by RND can generate the feasible plans for localization recovery notably in handling scenario changes and active initialization.

		\begin{table}[t]
			\begin{center}
				\caption{\textbf{Ablation experiment.} average navigation success rates of SCALE (the last row) and four ablations: not using affordance, using affordance without RND and RNN, using affordance without RND, and using affordance without RNN.}	
				\resizebox{1.0\linewidth}{!}{
					\begin{tabular}{cccc}
						\hline
						Methods &Garden &Office park &Plaza \\
						\hline
						SCALE w/o affordance &0.70 &0.75 &0.66 \\
						SCALE w/o RNN, RND &0.76 &0.80 &0.70 \\
						SCALE w/o RND &0.85 &0.86 &0.75 \\
						SCALE w/o RNN  &0.86 &0.86 &0.80 \\
						\rowcolor{Ocean} SCALE &\textbf{0.89} &\textbf{0.92} &\textbf{0.83} \\
						\hline
					\end{tabular}
				}
				\label{ablation2}
				\vspace{-0.5cm}
			\end{center}
			
		\end{table}
		
		

		\section{Conclusion}
		In this paper, we have proposed SCALE, a self-correcting visual navigation framework that can notably address the prevalent OOD issue. Our approach leverages IQL algorithm to learn an image-goal navigation policy from offline dataset. Furthermore, we propose a self-supervised localization recovery method that envisions future trajectories in OOD regions. Then, we employ the RND technique to learn a novelty estimator for evaluating and choosing the candidates that can guide the robot to recover the localization. By performing the experiments in three outdoor scenarios, we show that SCALE have highly strong robustness in handling the challenging real-world navigation tasks, surpassing the existing state-of-the-art methods and reducing the need for human intervention. In the future, we will extend our work to handle more challenging or unknown environments.
		\addtolength{\textheight}{-2cm}   

	\end{document}